\documentclass[10pt,twocolumn,letterpaper]{article}

\usepackage{iccv}
\usepackage{times}
\usepackage{epsfig}
\usepackage{graphicx}
\usepackage{amsmath}
\usepackage{amssymb}
\usepackage{soul}
\usepackage{comment}

\usepackage{algorithm}
\usepackage{subfig}
\usepackage{booktabs}
\usepackage{multicol}
\usepackage{multirow}
\usepackage{amsmath}
\usepackage[noend]{algpseudocode}
\usepackage[shortlabels]{enumitem}
\usepackage[]{xcolor}

\usepackage[breaklinks=true,bookmarks=false]{hyperref}

\iccvfinalcopy % *** Uncomment this line for the final submission

\def\iccvPaperID{5839} % *** Enter the ICCV Paper ID here

\ificcvfinal\fi

\newcommand{\para}[1]{{\noindent\textbf{#1}}}
\newcommand{\LTR}{\texttt{Learn-to-Race}}
\newcommand{\ltr}{\texttt{L2R}}

\begin{document}

%%%%%%%%% TITLE
%\title{Learn-to-Race: A Multimodal and \\Continuous Control Environment for Autonomous Racing}
\title{Learn-to-Race: A Multimodal Control Environment for Autonomous Racing}

\author{James Herman$^{1}$\thanks{Equal contribution.} \quad Jonathan Francis$^{1,2*}$ \quad Siddha Ganju$^{3}$ \quad Bingqing Chen$^{1}$ \quad Anirudh Koul$^{4}$\\ Abhinav Gupta$^{1}$ \quad Alexey Skabelkin$^{5}$ \quad Ivan Zhukov$^{5}$ \quad Max Kumskoy$^{5}$ \quad Eric Nyberg$^{1}$ \\
$^{1}$School of Computer Science, Carnegie Mellon University, Pittsburgh, PA, USA \\
$^{2}$Human-Machine Collaboration, Bosch Research, Pittsburgh, PA, USA \\
$^{3}$ NVIDIA, Santa Clara, CA, USA \\
$^{4}$ Pinterest, San Francisco, CA, USA \\
$^{5}$ Autonomous Driving, Arrival, London, UK \\
{\tt\small \{jamesher, jmf1, bingqinc, agupta6, ehn\}@cs.cmu.edu, \{sganju1, akoul\}@alumni.cmu.edu,} \\ {\small\tt \{skabelkin, zhukov, kumskoy\}@arrival.com}
}

\maketitle
\ificcvfinal\thispagestyle{empty}\fi
%the advancement of more sophisticated agents
%%%%%%%%% ABSTRACT
\begin{abstract}
%Existing research on autonomous driving primarily focuses on urban driving scenarios, leaving high-fidelity benchmarks for autonomous racing less explored. 

%Racing, compared to urban driving, is a challenging task that requires real-time, safety-critical inference in a highly dynamic environment. Existing simulators for autonomous racing are lacking in realism, with respect to visual rendering and vehicle dynamics, inhibiting the transfer of learning agents to real-world contexts. 

Existing research on autonomous driving primarily focuses on urban driving, which is insufficient for characterising the complex driving behaviour underlying high-speed racing. At the same time, existing racing simulation frameworks struggle in capturing realism, with respect to visual rendering, vehicular dynamics, and task objectives, inhibiting the transfer of learning agents to real-world contexts. We introduce a new environment, where agents \LTR~(\ltr) in simulated competition-style racing, using multimodal information|from virtual cameras to a comprehensive array of inertial measurement sensors. Our environment, which includes a simulator and an interfacing training framework, accurately models vehicle dynamics and racing conditions. In this paper, we release the Arrival simulator for autonomous racing. Next, we propose the \ltr~task with challenging metrics, inspired by learning-to-drive challenges, Formula-style racing, and multimodal trajectory prediction for autonomous driving. Additionally, we provide the \ltr~framework suite, facilitating simulated racing on high-precision models of real-world tracks. Finally, we provide an official \ltr~task dataset of expert demonstrations, as well as a series of baseline experiments and reference implementations. We make all code available: \url{https://github.com/learn-to-race/l2r}.

\end{abstract}

\begin{figure}
    \centering
    \includegraphics[width=0.235\textwidth]{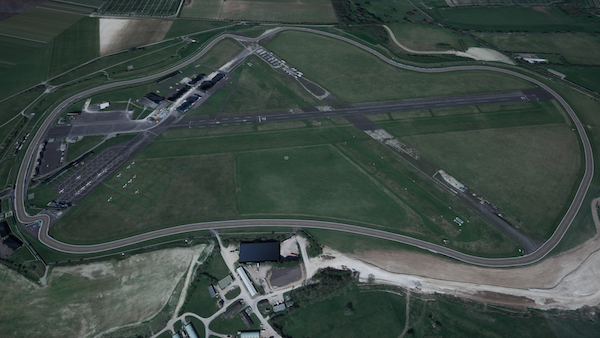}\hfill
    \includegraphics[width=0.235\textwidth]{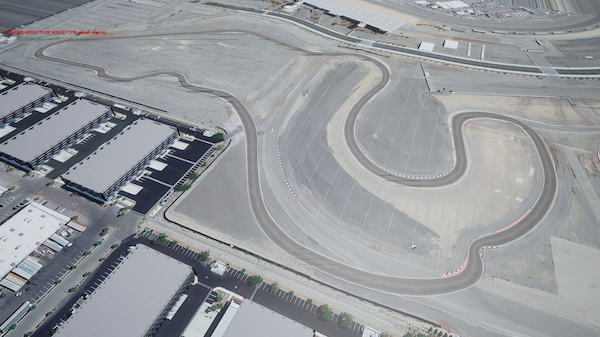}\hfill
    \\[\smallskipamount]
    \includegraphics[width=0.235\textwidth]{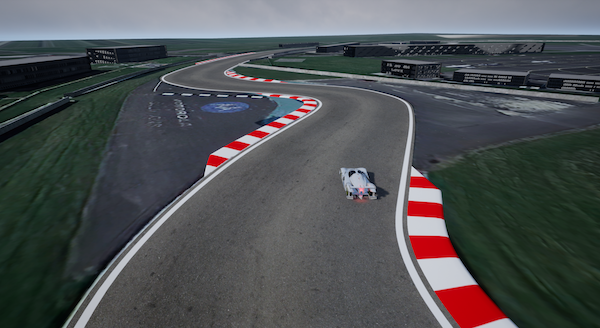}\hfill
    \includegraphics[width=0.235\textwidth]{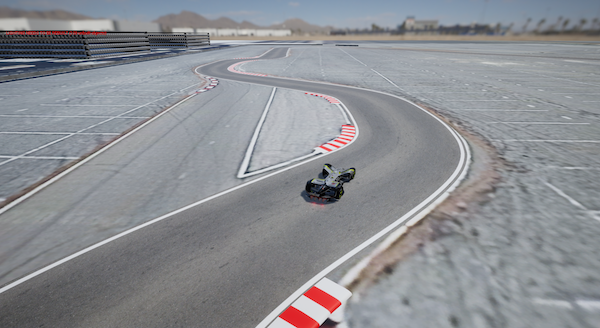}\hfill
    \\[\smallskipamount]
    \includegraphics[width=0.478\textwidth]{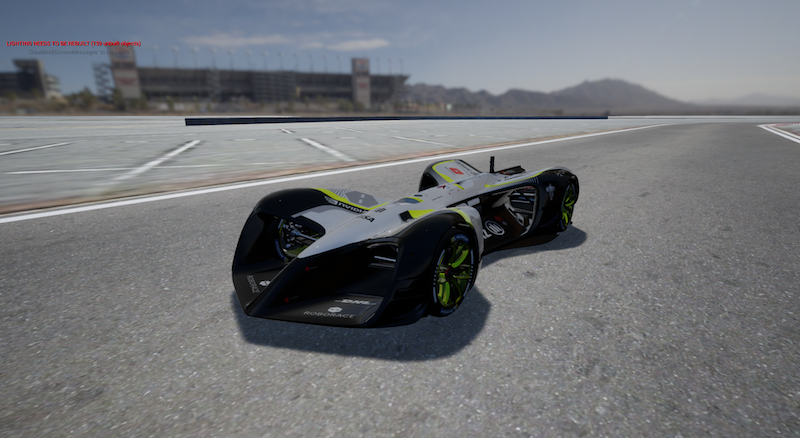}\hfill
    \caption{\small\LTR~interfaces with a racing simulator, which features numerous real-world racetracks such as the Thruxton Circuit (\textit{top-left}) and Las Vegas Motor Speedway (\textit{top-right}). Simulated race cars (\textit{bottom}) are empowered with learning agents, tasked with the challenge of learning to race for the fastest lap-times and best metrics.}
    \label{fig:simulator_images}
\end{figure}

%%%%%%%%% BODY TEXT
\section{Introduction}

Progress in the field of autonomous driving relies on the existence of challenging tasks and well-defined evaluation metrics, which enable researchers to effectively assess and improve algorithms. Models developed in learning-to-drive settings continue to struggle with issues in sample-complexity, safety, and unseen generalisation, calling for more suitable benchmarks \cite{chen2020learning, filos2020can, park2020diverse}. We hypothesise that high-fidelity simulation environments, together with well-defined metrics and evaluation procedures, are conducive to developing more sophisticated agents; and, in turn, such agents will be better-suited to real-world deployment.

%the presence of appropriately challenging tasks is conducive to the development of more sophisticated agents.
%Few simulation environments contain the complexity for assessing approaches on these bases and, similarly, few tasks are available that support metrics and evaluation procedures that are grounded in practical applications, while also being illustrative of the complexities of real-world deployment. 
%as much on the development of challenging tasks and suitable evaluation metrics as it does on the development of novel algorithms and architectures.

% for example, models continue to struggle with sample-efficiency, code portability, flexible inference, interpretability of driving behaviours, and generalisation to unseen scenarios. %Few autonomous racing simulators contain the complexity for assessing approaches on these bases and, similarly, few autonomous racing tasks are available that support metrics and evaluation procedures that are grounded in practical applications, while also being illustrative of the complexities of real-world deployment. 
 
Simulated autonomous racing exhibits task complexity on several factors: (i) agents must perform real-time decision making, requiring computationally-efficient policy updates as well as robustness to latency; (ii) agents must be able to deal with realistic vehicle and environmental dynamics (whereas agents in less-realistic environments have been able to achieve super-human performance); (iii) agents must leverage more informative intrinsic reward schemes that enable replication of human-like driving behaviour, e.g., trading off safety and performance; and (iv) agents must use offline demonstrations effectively, without overfitting, and must leverage interactions with the environment sample-efficiently. We highlight simulated racing (Figure \ref{fig:simulator_images}) as an opportunity for developing learning strategies that are capable of meeting these stringent requirements.

In this work, we release the Arrival Autonomous Racing Simulator, which includes numerous interfaces for both simulated and real vehicle instrumentation. Furthermore, we introduce \LTR~(\ltr), a multimodal and continuous control environment for training and evaluating autonomous racing agents. Through the \ltr~environment, we simulate competition-style racetracks that are based off real-world counterparts, we provide mechanisms for fully-characterising realistic racing agents (e.g., flexible sensor placements, multimodal cameras, and various vehicle dynamics profiles), and we provide numerous tools for fine-grained agent evaluation (e.g., random and fixed spawn locations, custom racing map construction, and injection of external disturbances). Using these facilities, we enable research in problems that require agents to make safety-critical, sub-second decisions in dynamically unstable contexts, such as autonomous racing, real-time uncertainty analysis in highway driving, and trajectory forecasting. In this paper, we exemplify algorithm development and benchmarking of methods under learning from demonstrations, reinforcement learning, and model-predictive control.

Concretely, our contributions include: (i) the Arrival Autonomous Racing Simulator, which simulates high-fidelity competition-style tracks, vehicles, and various sensor signals; (ii) the \LTR~(\ltr) framework, a plug-and-play environment, which defines interfaces for various sensor modalities and provides an OpenAI-gym compliant training and testing environment for learning-based agents; (iii) an official \ltr~task and dataset with expert demonstrations, metrics, and reference evaluation procedures; and (iv) an academic release of the simulator, code for the \ltr~framework, and implementations of baseline agents to facilitate full reproducibility and extension.

\begin{figure*}
\begin{center}
    \includegraphics[width=0.80\linewidth]{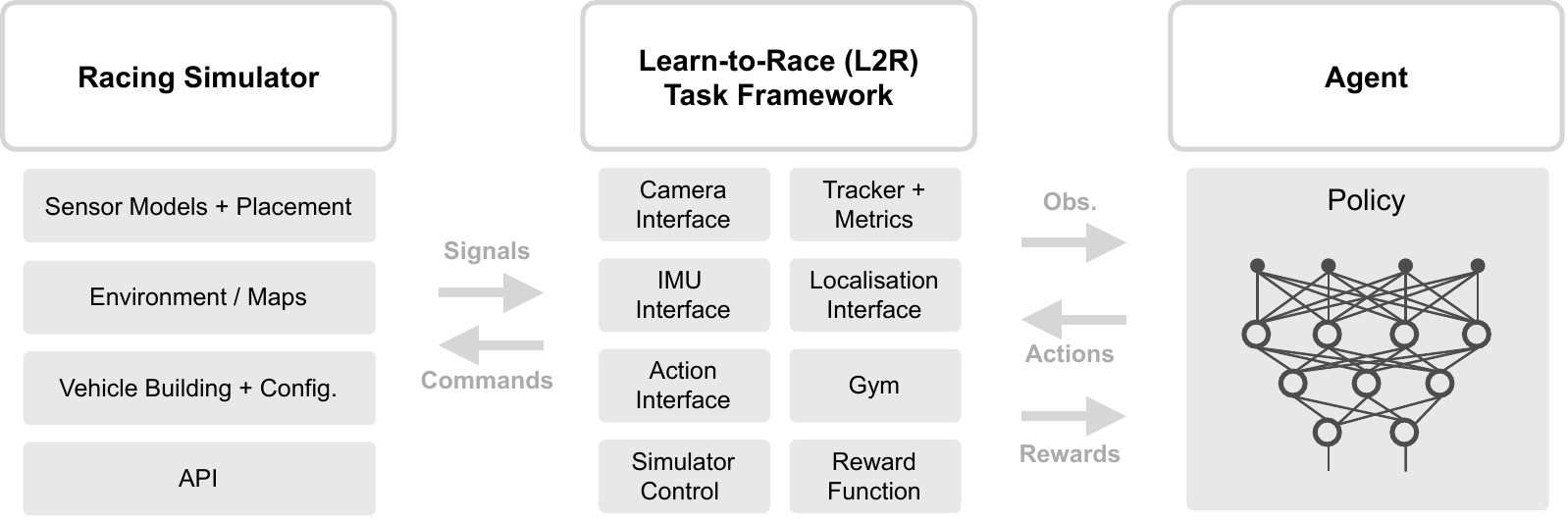}
\end{center}
   \captionsetup{margin=2.5cm}
   \caption{\small\LTR~allows agents to interact with the racing simulator through a series of interfaces for observations, actions, and simulator control.}
\label{fig:env_overview}
\end{figure*}
% link to edit: https://drive.google.com/file/d/1CH6NKfuVjvBMER7DcGbSM9mgCv9Ju2Ou/view?usp=sharing

%------------------------------------------------------------------------
\section{Related Work}

%-------------------------------------------------------------------------
\subsection{Reinforcement Learning Environments}

Research progress in the fields of Reinforcement Learning (RL), Planning, and Control has relied on various simulation environments, for benchmarking agent performance on game-playing and robot control tasks \cite{tassa2018deepmind, brockman2016openai, Todorov_mujoco:a, francis2021core, shao2019survey}. These tasks require sequential decision-making in order to complete objectives and are generally characterised by their state dimensionality, the nature of their action space (e.g., discrete or continuous), agent cardinality (i.e., single- or multi-agent), and by the capability of the underlying simulator in capturing real-world physical dynamics \cite{francis2021core}. Whereas the vast majority of tasks offered by, e.g., the DeepMind Control Suite \cite{tassa2018deepmind}, OpenAI Gym \cite{brockman2016openai}, and the MuJoCo physics engine \cite{Todorov_mujoco:a} have been solved|with agents often achieving superhuman performance|no existing environments focus on high-fidelity simulation of high-speed driving, in dynamically unstable contexts.

%Most involve either game playing or robotics control tasks, both of which require sequential decision-making to complete objectives. Game-based environments and competitions are well described by Shao et al. \cite{shao2019survey} which classifies games by state dimensionality and by whether there are single or multiple agents. A vast majority of these have been solved with agents achieving superhuman performance. Aside from games, DeepMind Control Suite \cite{tassa2018deepmind} and OpenAI Gym \cite{brockman2016openai} include many classical control and robotics tasks, some of which are powered by the MuJoCo physics engine \cite{Todorov_mujoco:a}. %Existing autonomous driving and racing simulators are described in more details in Section \ref{sec:review_racing}.

%-------------------------------------------------------------------------
% \subsection{Roborace}

%-------------------------------------------------------------------------
\subsection{Simulation of Autonomous Driving}\label{sec:review_racing}

\para{Urban driving.} CARLA \cite{dosovitskiy17carla} is an open-source simulator for autonomous driving, wherein various tasks have been defined to challenge agents' street-legal urban driving abilities. Duckietown \cite{duckietown, gym_duckietown} provides a customisable platform for urban autonomous driving, as well as hardware support for miniature vehicles controlled via RaspberryPi's. In this paper, we focus primarily on autonomous racing environments, which present challenges outside the conventional scope of urban and highway driving. %The OpenAI-gym compliant Gym-Duckietown \cite{gym_duckietown} is a customisable simulator for the Duckietown Universe. %Compared to urban driving, simulated competition-style...  Here, driving policies must make real-time decisions, for controlling vehicles near their physical limits, where even small actions have amplified control ramifications.

\para{Track racing.} In autonomous racing, agents must make sub-second decisions in regimes of unstable physical dynamics, wherein the ramifications of control actions can be amplified or suppressed, non-linearly, depending on vehicular and environmental state. CarRacing-v0, an OpenAI-gym environment \cite{brockman2016openai}, is a simple racing environment, which uses only bird's-eye-view (BEV) observations. In \cite{fuchs2020superhuman}, researchers trained agents to race in the video game Gran Turismo Sport, but have not yet released their environment. Moreover, instead of using sensory perception, agents were directly provided with privileged information, e.g., distance to obstacles and road boundaries. TORCS \cite{TORCS} is an open-source simulator and is used by the Simulated Car Racing Championship \cite{loiacono2013simulated}, despite its game-like qualities. As the goal of simulators should be to accurately model the dynamics of the real-world, we assert that the potential for model transfer from these frameworks remains limited.

\subsection{Learning Paradigms}

\noindent We discuss various learning paradigms that are enabled by the simulation of autonomous driving.\\

\para{Simulation-to-real transfer.} DeepRacer \cite{balaji2019deepracer}, developed by Amazon Web Services, provides an end-to-end framework for training and deploying 1/18th-scale autonomous racing cars. The Indy Autonomous Challenge \cite{indyac} encourages institutions to create autonomous vehicle technology; participants are given the proprietary VRXPERIENCE driving simulator, which focuses more on optimising human-machine interactions within the vehicle, in the context of situational highway driving, which contrasts with our focus in this work on autonomous racing. Roborace \cite{roborace} is the first international championship for full-size, real-world autonomous racing. Here, teams develop self-driving software and compete in challenges, using Roborace-owned vehicles. Roborace provides teams with proprietary software-in-the-loop (SIL) and hardware-in-the-loop (HIL) simulators, with a base driving stack. These simulators are predominately used for developing classical control methods, however, and do not include facilities for training learning-based agents \cite{betz17, Herrmann_2020, Herrmann_Wischnewski_2020, stahl2019}. The authors obtained a podium finish at the Thruxton Circuit (UK), in the Season Beta Roborace competitions (2020-2021), and we now wish to enable new technologies through open-sourcing our autonomous racing research: to our knowledge, we publicly release the first environment that is specifically intended for simulating autonomous competition-style track racing and for transferring learning-based agents to the real world.

\para{Safe and efficient learning.} Imposing safety constraints in, e.g., RL algorithms, has become popular for the potential of reducing failures in simulation-to-real transfer settings and for enabling agent robustness to environmental stochasticity \cite{garcia2015comprehensive}. The goal is to embed safety guarantees in policies, without compromising their performance or sample-efficiency. While a few works consider detection and avoidance of unsafe states, in urban driving \cite{ding2021multimodal} and human-assistive robotics \cite{fridovich2020confidence}, no existing works focus on safe learning and control, for autonomous racing in dynamically unstable contexts. Popular Safe-RL benchmarks (e.g., OpenAI Safety Gym) lack realistic dynamics and they evaluate agents at much lower speeds; thus, the numerous limitations of existing methods cannot be studied comprehensively. We assert that the physical realism that \ltr~provides facilitates improvement of those underlying approaches.

%Facilitates cheaper+safer real-world deployment, in sim-to-real transfer. We want to encourage the development of algorithms that are both computational- and sample-efficient; performing real-time decision making, requiring computationally-efficient policy updates as well as robustness to latency.

%Ansys, one of the sponsors of the Challenge, provides participants with access to their VRXPERIENCE Driving Simulator, and teams will have access to on-track testing prior to the event which takes place in October 2021 at the Indianapolis Motor Speedway.

%  The Roborace vehicle, Robocar, achieved a Guinness World Record in 2019 for the fastest autonomous car with a speed of 282.42 km/h (175.49 miles/h). 

%------------------------------------------------------------------------
\section{Simulation Environment}
\label{ssec:simulator}

% We introduce a learning to race framework that builds upon an existing racing simulator (Section 3.1). We create an interface to pass multimodal information between the simulator and an agent, enabling blah blah blah (Section 3.2).

\subsection{Arrival Autonomous Racing Simulator}

The Arrival simulator is a powerful tool for the development and testing of autonomous vehicles. It is based on Unreal Engine 4 and includes such features as: (i) a vehicle prototyping framework; (ii) full software-in-the-loop (SIL) simulation, to model all vehicle control devices; (iii) controller area network (CAN) bus interface; (iv) camera, inertial measurement unit (IMU), light detection and ranging (LiDAR), ultrasonic, and radar sensor models; (v) semantic segmentation; (vi) sensor placement and configuration facilities; (vii) V2V/V2I interface subsystem; (vii) dynamic racing scenario creation; (viii) race track generation from scanned datasets; (ix) support for full integration with the CARLA simulator \cite{dosovitskiy17carla}; and (x) an application programming interface (API), which is automatically generated based on C++ code analysis. Details in supplementary.

%------------------------------------------------------------------------
%\subsection{Simulator Parameters}

%\jh{coordinate system, list of all the interfaces, all parameters, place in appendix .. what can the simulator do?}

%------------------------------------------------------------------------
\subsection{Learn-to-Race Environment}
\LTR~(\ltr) is a multimodal control environment that provides a series of interfaces for an agent to interact with the racing simulator, including the capabilities to send control commands and make observations of the environment and vehicle state via different sensors (Figure~\ref{fig:env_overview}). \ltr~is implemented as a Gym environment \cite{brockman2016openai}, enabling quick prototyping of control policies. While we release the \ltr~environment and task (Section \ref{sec:task}) alongside the Arrival simulator, we note that other simulators may be used with \ltr, including those provided by \cite{roborace}. %Figure~\ref{fig:env_overview} shows a summary of functionalities and interactions amongst the racing simulator, the \ltr~framework, and an agent.\\

\para{Agent-Simulator Interaction.} At each step $t$, an agent selects an action $a_t$, based on its current observation $s_t$, using its policy $\pi_\theta$: $a_t\sim \pi_\theta(\cdot|s_t)$. The control action from the agent is forwarded to the simulator as a UDP message. \ltr~receives updates from the simulator, i.e., images from the virtual camera and/or measurements from other vehicle sensors, through TCP and UDP socket connections. As in reality, update frequencies across the various sensor modalities are not equal, so \ltr~synchronizes observations by providing agents with the most recent data from each (Algorithm \ref{step_algo}). The \texttt{step} method of the environment returns the new observation $s_{t+1}$, along with a calculated reward to the agent, $r_{t} = R(s_{t}, a_{t}, s_{t+1})$, and a Boolean terminal state flag. The reward function and evaluation metrics are defined in Section \ref{ssec:metrics}.

% \jf{Consider extending this part, especially with regard to the protocols; show the reader the complexity}

\begin{algorithm}
    \caption{Agent-Simulator Interaction}\label{}
    \begin{algorithmic}[1]
        \Function{Sensor Thread}{}
            \State $data \gets $ Initial value
            \Function{Get Data}{}
                \State return $data$
            \EndFunction
            \While{$not\:terminated$}
                \State $data \gets $ \textbf{Receive Data}
            \EndWhile
        \EndFunction
        \State
        
        \Function{Step}{$a_{t}$}
            \State \textbf{Send} $a_{t}$ as UDP message
            \State $s_{t+1} \gets $ \textbf{Get Data} $\forall$ \textbf{Sensor Threads}
            \State $r_{t} \gets R(s_{t}, a_{t}, s_{t+1})$
            \State $done \gets $ \textbf{IsTerminal($s_{t}, s_{t+1}$)}
            \State\Return $s_{t+1},\:r_{t},\:done$
        \EndFunction
    \end{algorithmic}
    \label{step_algo}
\end{algorithm}

\para{Episodic control.} The control interface communicates with the simulator to automatically setup and execute simulations in an episodic manner. \ltr~conveniently allows for training to be launched in one command, as all aspects of the racing simulator and learning environment are parameterised. A state is considered terminal if all laps are successfully completed, if at least 2 of the vehicle's wheels go outside of the drivable area, or if progress is minimally insufficient. The episode begins by resetting the vehicle to a standing start position, at a parameterised location along with configured sensor interfaces and initialised reward function. Discrete steps are taken by the agent until one of the aforementioned episode termination criteria is met. 

% While the simulator provides coordinates in GCS format, \ltr~converts position of the vehicle to the East, North, Up (ENU) coordinate system. The ENU system is a local coordinate system used to approximate the Earth's curvature to a flat surface tangential to the Earth. The primary benefit of is that the location expressed in ENU coordinate is easier for humans to interpret.

\begin{comment}
\subsection{Limitations}
Ultimately, we are motivated to bring agents which have learned to race to the real-world to compete in racing tasks with physical vehicles. The most significant challenge in doing so is the transfer from a simulated environment to the real-world. Since assumptions and simplifications are used to create simulated environments, including the introduced racing simulator, the bit accuracy of simulated hardware and race tracks is, to a significant degree, lower than their physical counterparts. Additionally, simulation-to-real transfer also requires seamless transfer from a hardware perspective. In this work, we consider such transfer to be out-of-scope.
\end{comment}
%\siddha{describe the possibility of Converting to Real World Car Racing, conversion work, bit accuracy, challenges}

%------------------------------------------------------------------------
\section{Task: Learn-to-Race}
\label{sec:task}

The \LTR~(\ltr) task tests an agent's ability to execute the requisite behaviours for competition-style track racing, through multimodal perceptual input. In this section, we provide a task overview and describe task properties, dataset characteristics, and metrics.

%------------------------------------------------------------------------
\subsection{Task Overview}

\ltr~provides an OpenAI Gym \cite{brockman2016openai} compliant learning environment, where researchers could flexibly select among the available sensor modalities. This early version of the environment enables single-agent racing on three racetracks (with custom track construction facility), modeled after their real-world counterparts. Included tracks are the Thruxton Circuit (\texttt{Track01:Thruxton}) and Anglesey National Circuit (\texttt{Track02:Anglesey}) from the United Kingdom, and the North Road Track at Las Vegas Motor Speedway (\texttt{Track03:Vegas}), located in the United States. Analogous to having separate town maps for training and testing in other simulation environments, e.g., CARLA \cite{dosovitskiy17carla}, we use \texttt{Track01} and \texttt{Track02} for training and \texttt{Track03} for testing. Consequently, we generate expert traces from the training tracks, for inclusion in our initial dataset release (see Section \ref{ssec:dataset}). Many avenues for research can be explored within \ltr, including various learning paradigms, such as: (constrained) reinforcement learning, learning from demonstrations, multitask learning, transfer learning and domain adaptation, simulation-to-real transfer, fast decision-making, classical/neural hybrid modeling, etc. Regardless of the method chosen, agents' multimodal perception capabilities|i.e., their ability to fuse and align sensory information|are of critical importance. 

%This first version of the environment provides access to two racetracks, both modeled after their real-world counterparts. The first is the Thruxton Circuit track (\texttt{Track01:Thruxton}), located in the United Kingdom, and the second is the North Road Track at Las Vegas Motor Speedway (\texttt{Track02:Vegas}), located in the United States.

%Of critical importance to the success of an agent is its multimodal perception capabilities: the generation of visual features via image encoding, image segmentation, visual attention, and the fusion of this context with that of other sensory input combinations.

\begin{table*}[t]
\small
  \centering
  \captionsetup{margin=1.7cm}
    \caption{\small Summary of the observation and continuous action spaces, for the \LTR~task. When the simulator is initialised in \textit{vision-only} mode, the observation space consists of just the images from the ego-vehicle's front-facing camera. The additional observation data, all of which is realistically accessible on a real racing car, is available in \textit{multimodal} mode. *Whereas gear is permitted as a controllable parameter, we do not use it in our experiments.}
  \label{table/:act_obs_space}
  \begin{tabular}{ p{2cm}p{4cm}p{6cm}p{1.5cm} }
    \toprule
     & \textbf{Signal} & \textbf{Description} & \textbf{Dimension} \\ [1pt]
    \midrule
    \multirow{2}{3cm}{\textbf{Action}} & Acceleration & Command in [-1.0, 1.0] & $\mathbb{R}^1 $  \\
    & Steering & Command in [-1.0, 1.0] & $\mathbb{R}^1$ \\ [1pt]
    & Gear & \{park, drive, neutral, reverse\} & |\\ [1pt]
    \midrule
    \multirow{13}{3em}{\textbf{Observation}} & Image & RGB image & $\mathbb{R}^ {W\times H\times 3}$\\ [1pt]
    %\cline{2-3}
    \\ [-10pt] % hacky
    & Steering & Observed steering direction&$\mathbb{R}^1 $ \\ [1pt]
    & Gear & \{park, drive, neutral, reverse\} & |\\ [1pt]
    & Mode & Vehicle mode &$\mathbb{R}^1 $ \\ [1pt]
    & Velocity & In ENU coordinate ($m/s$) & $\mathbb{R}^{ 3}$\\ [1pt]
    & Acceleration & In ENU coordinate ($m/s^{2}$) & $\mathbb{R}^{3}$\\ [1pt]
    & Yaw, Pitch, Roll & Orientation of the car ($rad$) & $\mathbb{R}^{3}$\\ [1pt]
    & Angular Velocity & Rate of change of the orientation ($rad/s$) & $\mathbb{R}^{3}$ \\ [1pt]
    & Location& Location of the vehicle center in ENU ($m$) & $\mathbb{R}^{3}$ \\ [1pt]
    & Wheel Rotational Speed & \textit{per wheel}  (RPM)& $\mathbb{R}^{4}$  \\ [1pt]
    & Braking & Brake pressure  \textit{per wheel} ($Pa$) & $\mathbb{R}^{4}$\\ [1pt]
    & Torque & \textit{per wheel} ($N\dot m$) & $\mathbb{R}^{4}$\\ [1pt]
    \bottomrule
  \end{tabular}
\end{table*}

%------------------------------------------------------------------------
\subsection{Learn-to-Race Dataset}
\label{ssec:dataset}

We generate a rich, multimodal dataset of expert demonstrations from training tracks, \texttt{Track01:}\texttt{Thruxton} and \texttt{Track02:}\texttt{Anglesey}, in order to facilitate pre-training of agents via, e.g., imitation learning (IL). The \ltr~dataset contains multi-sensory input at a 100-millisecond resolution, in both the observation and action spaces. Depending on the selected simulator perception mode, agents have access to one (\textit{vision-only mode}) or all modalities (\textit{multimodal mode}). See Table \ref{table/:act_obs_space} for a complete list of available modalities. The action space is defined by continuous values for acceleration and steering, in the ranges $[-1.0, 1.0]$, where negative acceleration values will decelerate the vehicle to a halted position. Note that \textit{Gear} is a controllable action, but fixed to \textit{drive} in all our experiments. %Setting gear to park, neutral, or reverse does not help in the racing objectives. %\jf{Is vehicle mode part of the action space also, or would reversing just require negative acceleration values, once the vehicle's forward acceleration has diminished completely?} \jh{I removed gear from the action space, but the simulator does allow one to change gears} \jf{Got it; Or was there a strong reason why you think it should be omitted? Otherwise, we might as well include it, for the sake of completion.}

%For each racetrack, we provide image datasets from the vehicle's virtual camera which were collected by driving manually driving around the tracks in a zig-zag fashion to increase the diversity of the images.

The expert demonstrations were collected using a model predictive controller (MPC) (Section \ref{ssec:baselines}), which follows the centerline of the race track at a pre-specified reference speed. This first version of the dataset contains 10,600 samples of each sensor and action dimension, for 9 complete laps around each track. Future version releases of \ltr~will include access to new simulated tracks (modeled after other real tracks, around the world) as well as expert traces generated from these additional tracks|across various weather conditions, in challenging multi-agent settings, and within dangerous obstacle-avoidance scenarios. 

%\jf{The dataset should be as broad and versatile as possible; it should include all sensor modalities, position information, and actions executed at each time-step; it should support all the different types of (potential) model/agent training paradigms. In Section \ref{ssec:metrics}, we constrain the inputs to just those we wish to allow for the ``official" task, but we don't want to constrain the more-general research that can be done on the dataset / environment / simulator, outside of our defined task|people will need to cite us, either way.}

%\begin{figure}%
%    \centering
%    \subfloat[]{\includegraphics[width=0.8\linewidth]{assets/sample_image_thruxton_narrow.png}
%    }\\
%    \subfloat[]{\includegraphics[width=0.8\linewidth]{assets/sample_image_lvms_narrow.png}
%    }
%    \caption{\small Frontal-camera views on (a) \texttt{Track01:Thruxton} and (b) \texttt{Track03:Vegas}, sampled from expert demonstrations.}% %Sample images from our \ltr~ datasets: (a) Thruxton Circuit, in the United Kingdom, which we use as the \ltr~task training track; and (b) North Road Track, at the Las Vegas Motor Speedway, selected as the test track.}% % emphasise the visual differences, between the two tracks
%    \label{fig:example}%
%\end{figure}

%------------------------------------------------------------------------
\subsection{Task Metrics}
\label{ssec:metrics}

%\jf{Talk about the official modalities, metrics, and evaluation strategies we decided to define for the \ltr~task.}

%There are many avenues that can be explored in trying to complete this task including imitation learning and reinforcement learning. Even more vibrant are the possibilities of how to perceive the world through the multiple input modalities available to agents. Of critical importance is the computer vision capabilities of agents; the generation of visual features via image encoding, segmentation, attention, some combination of these, or through novel approaches may significantly impact overall performance. \jh{Classical control solutions, [no planning, predetermined trajectories, how to word this appropriately?] are not consistent with the task of learning to race.}

The primary objective of the \ltr~task is to minimise the time taken for an agent to successfully complete racing laps, with additional requirements on the agent's driving quality. We do not restrict the agent's learning paradigms to, e.g., IL or RL; on the contrary, we can envision a wealth of combination strategies and other methods that are applicable to the task. While we do not include planning-\textit{only} approaches as those that are consistent with the official \ltr~task, (i) we do encourage hybrid or model-based learning approaches; furthermore, (ii) we do encourage the simulator and the \ltr~interface to be used to further research in these areas, more generally. Agnostic to the learning paradigm used, and inspired by concepts from high-speed driving, robot navigation \cite{francis2021core}, and trajectory forecasting \cite{park2020diverse}, we define the core modalities, metrics, and objectives that shall be used to train \ltr~agents and assess their performance. We summarise agents' action and observation spaces in Table \ref{table/:act_obs_space} and the official \ltr~task metrics in Table \ref{table/:metrics}.

We define the successful completion of an episode in the \ltr~task to be 3 completed laps, from a standing start; \textit{Episode Completion Percentage} (ECP) measures the amount of the episode completed, and \textit{Episode Duration} (ED) measures the minimum amount of time that the agent took to progress to its furthest extent, through the episode. We define \textit{Average Adjusted Track Speed} (AATS) as a metric that measures the average speed of the agent, across all three laps of the episode. Metrics may also include adjustments for environmental factors, such as wheel slippage and weather effects as the task matures. \textit{Average Displacement Error} (ADE), a common metric in trajectory forecasting \cite{park2020diverse}, measures the agent's average deviation from a reference path|in this case, the centerline of the track. \textit{Trajectory Admissibility} (TrA) is the dimensionless metric $\alpha$, defined in Equation \ref{eq:traj_admissibility}, where $t_{e}$ is the duration of the episode and $t_{u}$ is the cumulative time spent driving unsafely with exactly one wheel outside of the drivable area.
\begin{equation}
    % Trajectory Admissibility
    \alpha = 1 - \sqrt{\frac{t_{\text{u}}}{t_{e}}}
    \label{eq:traj_admissibility}
\end{equation}
\noindent We also utilise metrics that measure the smoothness of agent behaviour: \textit{Trajectory Efficiency} (TrE) measures the ratio of track curvature to agent trajectory curvature, i.e., in terms of agent heading deviations; \textit{Movement Smoothness} (MS) quantifies the smoothness of the agent's acceleration profile, adjusted for gravity, using the negated log dimensionless jerk, $\eta_{ldj}$ in Equation \ref{eq:smoothness}, inspired by \cite{quantifying_smoothness}:
\begin{equation}
    % Movement smoothness
    % Log dimensionless jerk
    % \cite{quantifying_smoothness}
    \eta_{ldj} = \ln \left( \frac{(t_{2}-t_{1})^{3}}{v_{peak}^{2}}
    \int_{t_{1}}^{t_{2}}\left\lvert\frac{d^{2}v}{dt^{2}}\right\rvert^{2} dt \right)
    \label{eq:smoothness}
\end{equation}
%One key property of \ltr~is that it is objective-centric. 
Rather than restricting agents to predefined incentive policies, input dimensions, or even input modalities, \ltr~allows and encourages flexibility so that agents can learn to race effectively. The default reward function for \ltr~is inspired by \cite{fuchs2020superhuman}: this policy provides dense rewards for progression, consistent with the goal of minimising lap times, and negative rewards for going out-of-bounds.

\subsection{Task Evaluation Procedure}
\label{ssec:task_evaluation}

Agent assessment is conducted through a leaderboard competition, with two distinct stages: (1) pre-evaluation and (2) evaluation. Predicated on industry standards, we adopt a racing-centric pre-evaluation step, for assessing agent performance, giving agents a warm start on the test track before formal evaluation. Much like how human racing drivers are permitted to acquaint themselves with a new racing track, before competition, we run a pre-evaluation on models, with unfrozen weights, allowing for some initial (albeit constrained) exploration. In this pre-evaluation period, agents may %one of the following options: (1) complete a single lap around the test track successfully or unsuccessfully; or (2) 
explore the environment for a fixed time of 60 minutes, defined in the number of time-steps of discrete observation from the \ltr~framework. In the pre-evaluation, we further define a ``competency check'' that agents must pass, in order to successfully proceed through to the main evaluation phase. For the North Road track at Las Vegas Motor Speedway, the only competency check is that agents are able to successfully complete a lap during the pre-evaluation period with acceleration capped at 50\% of maximum allowed in the action space. A successful episode is defined the completion of 3 laps from a standing start and the agent not going out of the driveable areas of the track. If the agent is unsuccessful in the pre-evaluation phase, it is disqualified and not evaluated further.  %  In the event that an agent does not score above a fixed threshold on the checks, the agent is unsuccessful and is not evaluated further. The user can submit agents for evaluation to the leaderboard, once daily. %Competency checks vary depending on the test track and ensure that the agent does not foray into the non-driveable area of the map. We define the following as competency checks for the North Road track at Las Vegas Motor Speedway, given its frequent, sharp turns and speed traps: (1) driving along the track centerline, with {\color{red}minimal} centerline deviation, (2) executing the challenging turn in the northeast section of the track, and (3) passing through a speed trap (sequential hairpin curves). %
As we continue to provide support for new tracks (necessitating more novel driving maneuvers), we will also continue to add and permute the driving competency checks, to maintain fairness of evaluation on those tracks.
%Regardless of which option is selected, %complete a fixed iterations of training to receive a fixed number of observations from the environment. 
%The agent can choose which option serves its model the best

Post a successful pre-evaluation stage, the final test stage occurs: agents are provided all the various input modalities and have to compete on the metrics defined Section \ref{ssec:metrics}. When the agent successfully passes through the pre-evaluation stage, the user is not provided with the results of the competency checks and instead is able to view the results of the complete evaluation directly on the leaderboard. %With multiple metrics available to evaluate on, the agent that performs the best on 

%and the score for the episode, $s_{e}$, to be the total time elapsed to complete. Failure to complete an episode results in an episode score, $s_{e}$, of $\infty$. More specifically, the benchmark metric is the total race time for the 5 best scores out of 6 consecutive episodes. This metric punishes, to some degree, agents that engage in risky racing behavior.

%\siddha{
%- trajectory 
%- unsafe conditions
%- latency
%- response time
%-}

\begin{table*}[h]
\small
  \centering
    \captionsetup{margin=1.4cm}
    \caption{\small\LTR~defines multiple metrics for the assessment of agent performance. These metrics measure overall success|e.g., whether and how fast the task is completed|along with more specific properties, such as trajectory admissibility and smoothness.}
  \label{table/:metrics}
  \begin{tabular}{ll}
    \toprule
    \textbf{Metric} & \textbf{Definition} \\
    \midrule
    \textit{Episode Completion Percentage} & Percentage of the 3-lap episode completed \\ [1pt]
    \textit{Episode Duration} & Duration of the episode, in seconds \\ [1pt]
    \textit{Average Adjusted Track Speed} & Average speed, across all three laps, adjusted for environmental conditions, in km/h \\ [1pt]
    \textit{Average Displacement Error} & Euclidean displacement from (unobserved) track centerline, in meters \\ [1pt]
    \textit{Trajectory Admissibility} & Complement of the square root of the proportion of cumulative time spent unsafe \\ [1pt]
    \textit{Trajectory Efficiency} & Ratio of track curvature to trajectory curvature (i.e., in agent heading) \\
    \textit{Movement Smoothness} & Log dimensionless jerk based on accelerometer data, adjusted for gravity \\
    %\textit{Braking Smoothness} & Inverse Euclidean displacement, between braking trajectory and Lipschitz smooth curve\\ [1pt]
    \bottomrule
  \end{tabular}
\end{table*}

\begin{table*}[!ht]
\small
  \centering
  \captionsetup{margin=1.7cm}
  \caption{\small Baseline agent results on \LTR~task while training on Thruxton track, with respect to the task metrics in Table \ref{table/:metrics}: Episode Completion Percentage (\textbf{ECP}), Episode Duration (\textbf{ED}), Average Adjusted Track Speed (\textbf{AATS}), Average Displacement Error (\textbf{ADE}), Trajectory Admissibility (\textbf{TrA}), Trajectory Efficiency (\textbf{TrE}), and Movement Smoothness (\textbf{MS}). Arrows ($\uparrow\downarrow$) indicate directions of better performance. Asterisks (*) in Tables \ref{table/:train-results} and \ref{table/:test-results} indicate metrics which may be misleading, for incomplete racing episodes.}
  \label{table/:train-results}
  \begin{tabular}{lccccccc}
    \toprule
    \textbf{Agent} & \textbf{ECP} ($\uparrow$) & \textbf{ED} ($\downarrow$) & \textbf{AATS} ($\uparrow$) & \textbf{ADE} ($\downarrow$) & \textbf{TrA} ($\uparrow$) & \textbf{TrE} ($\uparrow$) & \textbf{MS} ($\uparrow$) \\
    \midrule
    %\texttt{HUMAN[LV]} & $100.0(\pm 0.0)$ & $176.2(\pm 3.4)$ & $114.2(\pm 2.3)$ & $1.7(\pm 0.1)$ & $0.88(\pm 0.01)$ & $1.09(\pm 0.02)$ & $10.1(\pm 0.3)$ \\ [1pt]
    \texttt{HUMAN} & $100.0(\pm 0.0)$ & $235.8(\pm 1.7)$ & $171.2(\pm 3.4)$ & $2.4(\pm 0.1)$ & $0.93(\pm 0.01)$ & $1.00(\pm 0.02)$ & $11.7(\pm 0.1)$ \\ [1pt]
    \hline
    %\texttt{RANDOM[LV]} &  $1.0(\pm 0.6)$ & $21.9(\pm 9.6)$ & $9.2(\pm 1.5)$ & $1.4(\pm 0.3)$ & $0.74(\pm 0.01)$ & $0.18(\pm 0.05)^{*}$
    %& $8.4(\pm 1.0)$ \\ [1pt]
    \texttt{RANDOM} &  $0.5(\pm 0.3)$ & $14.0(\pm 5.5)$ & $11.9(\pm 3.8)$ & $1.5(\pm 0.6)$ & $0.81(\pm 0.04)$ & $0.33(\pm 0.38)^{*}$
    & $6.7(\pm 1.1)$ \\ [1pt]
    %\texttt{MPC[LV]} &  $69.5(\pm 10.7)$ & $353.2(\pm 54.8)$ & $40.5(\pm 0.1)$ & $0.8(\pm 0.1)$ & $0.91(\pm 0.02)$ & $1.07(\pm 0.01)^{*}$ 
    %& $10.4(\pm 0.2)$ \\ [1pt]
    \texttt{MPC} &  $100.0(\pm 0.0)$ & $904.2(\pm 0.7)$ & $45.1(\pm 0.0)$ & $0.9(\pm 0.1)$ & $0.98(\pm 0.01)$ & $0.85(\pm 0.03)$ & $10.4(\pm 0.6)$ \\ [1pt]
    %\hline
    %\texttt{RL-SAC[PE-LV]} &  $11.8(\pm 0.1)$ & $109.9(\pm 7.5)$ & $22.1(\pm 1.5)$ & $1.3(\pm 0.1)$ & $0.95(\pm 0.01)$ & $0.58(\pm 0.01)*$
    %& $9.9(\pm 0.2)$ \\ [1pt]
    %\texttt{RL-SAC[LV]} &  $88.1(\pm 16.8)$ & $290.2(\pm 50.3)$ & $61.8(\pm 1.4)$ & $1.7(\pm 0.1)$ & $0.92(\pm 0.02)$ & $0.47(\pm 0.01)^{*}$ 
    %& $12.6(\pm 0.3)$ \\ [1pt]
    \texttt{RL-SAC} &  $31.1(\pm 0.0)$ & $251.2(\pm 1.4)$ & $50.5(\pm 0.3)$ & $0.5(\pm 0.0)$ & $0.97(\pm 0.0)$ & $0.48(\pm 0.0)^{*}$ 
    & $11.1(\pm 0.4)$ \\ [1pt]
    %\texttt{IL-CIL} &  $0.00(\pm 0.0)$ & $00000(\pm 000)$ & $0.0(\pm 0.0)$ & $0.0(\pm 0.0)$ & $0.0(\pm 0.0)$ & $0.0(\pm 0.0)$ & $0.0(\pm 0.0)$ \\ [1pt]
    \bottomrule
  \end{tabular}
\end{table*}

\begin{table*}[!ht]
\small
  \centering
  \captionsetup{margin=1.7cm}
  \caption{\small Baseline agent results on \LTR~task while testing on Las Vegas track.}
  \label{table/:test-results}
  \begin{tabular}{lccccccc}
    \toprule
    \textbf{Agent} & \textbf{ECP} ($\uparrow$) & \textbf{ED} ($\downarrow$) & \textbf{AATS} ($\uparrow$) & \textbf{ADE} ($\downarrow$) & \textbf{TrA} ($\uparrow$) & \textbf{TrE} ($\uparrow$) & \textbf{MS} ($\uparrow$) \\
    \midrule
    \texttt{HUMAN} & $100.0(\pm 0.0)$ & $176.2(\pm 3.4)$ & $114.2(\pm 2.3)$ & $1.7(\pm 0.1)$ & $0.88(\pm 0.01)$ & $1.09(\pm 0.02)$ & $10.1(\pm 0.3)$ \\ [1pt]
    %\texttt{HUMAN[TH]} & $100.0(\pm 0.0)$ & $235.8(\pm 1.7)$ & $171.2(\pm 3.4)$ & $2.4(\pm 0.1)$ & $0.93(\pm 0.01)$ & $1.00(\pm 0.02)$ & $11.7(\pm 0.1)$ \\ [1pt]
    \hline
    \texttt{RANDOM} &  $1.0(\pm 0.6)$ & $21.9(\pm 9.6)$ & $9.2(\pm 1.5)$ & $1.4(\pm 0.3)$ & $0.74(\pm 0.01)$ & $0.18(\pm 0.05)^{*}$
    & $8.4(\pm 1.0)$ \\ [1pt]
    %\texttt{RANDOM[TH]} &  $0.5(\pm 0.3)$ & $14.0(\pm 5.5)$ & $11.9(\pm 3.8)$ & $1.5(\pm 0.6)$ & $0.81(\pm 0.04)$ & $0.33(\pm 0.38)^{*}$
    %& $6.7(\pm 1.1)$ \\ [1pt]
    \texttt{MPC} &  $69.5(\pm 10.7)$ & $353.2(\pm 54.8)$ & $40.5(\pm 0.1)$ & $0.8(\pm 0.1)$ & $0.91(\pm 0.02)$ & $1.07(\pm 0.01)^{*}$ 
    & $10.4(\pm 0.2)$ \\ [1pt]
    %\texttt{MPC[TH]} &  $100.0(\pm 0.0)$ & $904.2(\pm 0.7)$ & $45.1(\pm 0.0)$ & $0.9(\pm 0.1)$ & $0.98(\pm 0.01)$ & $0.85(\pm 0.03)$ & $10.4(\pm 0.6)$ \\ [1pt]
    %\hline
    \texttt{RL-SAC} &  $11.8(\pm 0.1)$ & $109.9(\pm 7.5)$ & $22.1(\pm 1.5)$ & $1.3(\pm 0.1)$ & $0.95(\pm 0.01)$ & $0.58(\pm 0.01)^{*}$
    & $9.9(\pm 0.2)$ \\ [1pt]
    %\texttt{RL-SAC} &  $88.1(\pm 16.8)$ & $290.2(\pm 50.3)$ & $61.8(\pm 1.4)$ & $1.7(\pm 0.1)$ & $0.92(\pm 0.02)$ & $0.47(\pm 0.01)^{*}$ 
    %& $12.6(\pm 0.3)$ \\ [1pt]
    %\texttt{RL-SAC[TH]} &  $31.1(\pm 0.0)$ & $251.2(\pm 1.4)$ & $50.5(\pm 0.3)$ & $0.5(\pm 0.0)$ & $0.97(\pm 0.0)$ & $0.48(\pm 0.0)^{*}$ 
    %& $11.1(\pm 0.4)$ \\ [1pt]
    %\texttt{IL-CIL} &  $0.00(\pm 0.0)$ & $00000(\pm 000)$ & $0.0(\pm 0.0)$ & $0.0(\pm 0.0)$ & $0.0(\pm 0.0)$ & $0.0(\pm 0.0)$ & $0.0(\pm 0.0)$ \\ [1pt]
    \bottomrule
  \end{tabular}
\end{table*}

\section{Baseline Agents}
\label{ssec:baselines}

We define a series of learning-free (e.g., \texttt{RANDOM}, \texttt{MPC}) and learning-based (e.g., reinforcement learning, imitation learning) baselines, to illustrate the performance of various algorithmic classes on the \ltr~task. We also benchmark human performance, through a series of expert trials. \\ %In addition to the RL and IL baselines, we also provide three additional baselines: , and \texttt{HUMAN}.\\
%\subsection{Reinforcement Learning}

\para{Random.} The \texttt{RANDOM} agent is mainly intended as a simple demonstration of how to interface with the \ltr~environment. The \texttt{RANDOM} agent is spawned at the start of the track, and uniformly samples actions, i.e., steering and acceleration, from the action space. The agent then proceeds to execute these random actions.

\para{MPC.} The MPC was used to generate expert demonstrations (Section \ref{ssec:dataset}) and is intended as a reference solution of \ltr~via classical control approaches. The MPC minimizes the tracking error with respect to the centerline of the race track at a pre-specified reference speed. We use the iterative linear quadratic regulator (iLQR) proposed in \cite{li2004iterative}, which iteratively linearizes the non-linear dynamics along the current estimate of trajectory, solves a linear quadratic regulator problem based on the linearized dynamics, and repeats the process until convergence. Specifically, we used the implementation for iLQR from \cite{amos2018differentiable}. We adopt the kinematic bike model \cite{kong2015kinematic} to characterize the vehicle dynamics. Further MPC details are provided in the supplementary.

While MPC implies optimal control performance, we want to point out the limitations of our current implementation. Firstly, the ground truth vehicle parameters were not known to us and we used estimated values. Secondly, we asked the MPC to follow the centerline of the track, which is not the trajectory expert drivers would have taken, especially when cornering. Finally, we pre-specified the MPC to drive at a conservative speed (12.5m/s), which makes the expert demonstrations easier to learn from.

%\subsection{Imitation Learning}
\para{Conditional Imitation Learning.} We adopted the same neural architecture from \cite{codevilla2018end}, except that we do not have different commands in our case, e.g., turn left, turn right, go straight, and stop. Thus, we used a single branch for decoding actions. We assume both front view images and sensor measurements are available for the IL agent. In each sample, the input consists of a $512\times 384$ image and 30 sensor measurements, and output is 2 actions (as listed in Table \ref{table/:act_obs_space}). The implementation of CIL automatically adjusts the neural network architecture based on specified input-output dimensions. The imitation loss (Equation \ref{eq:imit_loss}) is the mean squared error between the predicted action, $\hat{a}_t$, and the action taken by the expert, $a_t$.

\begin{equation}\label{eq:imit_loss}
\mathcal{L} = \sum_{i=1}^n||\hat{a}_i-a_i||^2_2
\end{equation}
%\subsection{Additional Baselines}

\para{Soft Actor-Critic.} We provide a reference implementation of Soft-Actor Critic (SAC) \cite{baselines, pmlr-v80-haarnoja18b}, which is generally performant and known to be robust \cite{eysenbach2021maximum}. SAC belongs to the family of maximum entropy reinforcement learning (RL) algorithms, wherein an agent maximizes expected return, subject to an entropy regularization term (Equation \ref{eq:loss_sac}), as a principled way to trade-off exploration and exploitation.
\begin{equation}\label{eq:loss_sac}
\mathcal{J}(\theta) = \sum_{t=1}^T \mathbb{E}_{\pi_\theta}[R(s_t, a_t)-\mathcal{H}(\pi_\theta(a_t|s_t))]
\end{equation}
Our \texttt{RL-SAC} agent demonstrates several of features in the environment: it operates in vision-only mode, but rather than learning directly from pixels, we pre-trained a convolutional, variational auto-encoder \cite{kingma2014autoencoding} made on sample camera images. Therefore, our agent only need to learns to decode actions from image embeddings using a multi-layer perceptron with two hidden layers of 64 hidden units each. Our agent's reward function was the environment's default with the inclusion of a bonus if the agent remained near the center of the track. % which along with restricting the action space, allowed the agent to successfully complete its first lap in fewer time steps at the expense of asymptotic performance.%\footnote{We will make our implementation code and checkpoints of this SAC-based agent available.

\para{Human.} We additionally establish a \texttt{HUMAN} performance baseline, by collecting simulated racing results from human expert players. The collection procedure involved a private crowd-sourcing event, which was split into two separate phases|practice/training and recording/testing. Expert players were already familiar with the simulator, task, and objective, prior to engaging in the event. In the training phase, players were instructed to engage in the race, until the variance in finished lap-times, for three consecutive runs, fell below a certain threshold. After this training phase, players were allowed to proceed to the testing phase, for which their top-3 laps were recorded. We averaged the top-3 results in the testing phase, from all experts, for each track; the training results were discarded.
%------------------------------------------------------------------------

\section{Experiments and Results}

We evaluate each of the baseline agents|\texttt{HUMAN}, \texttt{RANDOM}, \texttt{MPC}, and \texttt{RL-SAC}|on the \ltr~task, with the objective of finishing 3 consecutive laps in minimal time. For all approaches, agents complete model training and tuning on \texttt{Track01:Thruxton}. We present the average of each metric across 3 consecutive episodes, in Table \ref{table/:train-results}. Afterwards, agents are evaluated based on their performance on \texttt{Track03:Vegas}, following the 1-hour pre-evaluation period described in Section \ref{ssec:task_evaluation}. Learning-free agents, \texttt{RANDOM} and \texttt{MPC}, simply perform inference in the testing environment. The \texttt{RL-SAC} agent, a learning-based approach, operates in vision-only mode and utilizes the pre-evaluation stage to perform simple transfer learning to the new racetrack. The agent's image encoder does not have access to the test track prior to pre-evaluation and is not updated during this phase, but the model weights of the agent do update as new experience becomes available. Following the pre-evaluation phase, agents completed 3 consecutive episodes, and we present metric averages in Table \ref{table/:test-results}.

\para{Human experts.} Human experts strongly outperform during both training and testing, suggesting a general understanding of racing: they can quickly adapt to a new track, despite different features, including frequent and severe turns. Human experts fully complete 3 lap episodes at speeds near the vehicle's physical limits and estimate their lap-time performance to be within 10\% of optimal. We expect strong agents to execute trajectories which are of lower curvature than the racetrack's centerline, or a \textbf{TrE} of at least 1.0, allowing the vehicle to maintain higher \textbf{AATS}. Only human experts were able to achieve this, considering that failure to complete an episode distorts the metric. However, such trajectories are aggressive and risky, because they often involve cutting corners with the wheels nearly outside of the driveable area; this is apparent by higher \textbf{ADE} values and relatively low \textbf{TrA}. Additionally, human experts performed well relative to other agents terms of \textbf{MS}, demonstrating the ability to anticipate the need for acceleration and to apply smooth control.

\para{Baseline agents.} There are several notable conclusions that we make based on the performance of our baseline agents which we do not claim to be state-of-the-art. The first is that the task is indeed challenging, as even the \texttt{MPC} agent with an approximate car model failed to consistently complete laps on the test track. Even after over 1 million steps environment steps on the training track, the \texttt{RL-SAC} agent only completes about 90\% of a lap due to the challenging speed trap near the finish line at Thruxton. However, the \texttt{RL-SAC} agent demonstrates better control than the \texttt{MPC} in training in both \textbf{ADE} and \textbf{MS}. Second, we note the lack of generalization and poor sample efficiency of the \texttt{RL-SAC} agent whose performance dropped significantly in terms of ability to progress down the track, \textbf{ECP}, and stay near the centerline, \textbf{ADE}, despite being directly incentivised to do so. The agent learns to simply stop altogether to avoid going out-of-bounds about 1/3rd of the way around the test track. We note that imitation learning has potential for providing agents with strong priors. However, in our experiments, automatic network sizing based on input/output dimensions and step-wise supervision alone, suggested by \cite{dosovitskiy17carla}, did not yield good performance. This demonstrates the challenge that \ltr~poses to this family of approaches, necessitating consideration of, e.g., joint IL/RL strategies.

%-----------------------------------------------------------------------

\section{Discussion}

We are confident that agents can achieve superhuman performance for any given track given that (1) they are sufficiently complex and (2) that they have interacted with that environment enough times. What is not clear, is how well agents can generalize to new racetracks in a realistic simulation environment. We believe the \LTR~task will effectively assess models, based on their general understanding of vehicle dynamics, high-speed and high-risk control, racetrack perception, and intelligent racing tactics. 

To challenge state-of-the-art learning approaches, which continue to demonstrate superhuman performance in simplistic environments, we believe that the direction of future tasks must be towards higher complexity and realism. Our racing simulator has been used as a primary modeling tool for autonomous agents which have demonstrated real-world racing speeds in excess of 200 km/h, an order of magnitude faster, and more complex, than comparable environments. Limitations of our simulation environment with respect to competing simulators include multi-agent racing and a (currently) limited supply of tracks|however, both multi-agent racing and additional tracks will follow. Future enhancements also include additional vehicle sensors, domain randomization, and support for distributed training in learning-based approaches. We believe these enhancements serve as a precursor to real-world transfer and safety learning.

% - advantage of simulator and why yet another simulator: the direction for these simulators should be toward realism; current publicly-available simulators and racing frameworks do not attempt to capture vehicle and environmental dynamics at the granularity most appropriate for transfer to the the real-world. \\
% - why is the task important? reproducible, \\ - what is not covered by this work? \\ - adding new tracks to the task, adding more tasks \\ - what will happen in the future \\ - sponsor future research in area/potential impacts? \\
\iffalse
\begin{itemize}
    \item Reinforcing assumptions - racing on a new track is a basic task for a human, not so for learning-based approaches
    \item Double down on simulator capabilities, need for complex environments
    \item Limitations - 
    \item Future work - more tracks, new tasks (multi-agent, obstacle avoidance), enhancements to simulation environment (more sensors, weather, distributed training)
\end{itemize}
\fi

% Addressing computational practicalities, such as performing the same task in constrained resources

% The \ltr~framework provides a foundation for the development of a variety of racing tasks using a realistic simulation environment. We anticipate numerous improvements such as the introduction of new racetracks, simulation features, and racing tasks including multi-agent racing. Despite the racing simulator's advanced capabilities such as hardware integration, the task of simulation-to-real transfer remains out-of-scope.

%------------------------------------------------------------------------
\section{Conclusion}

We have presented: (i) a high-fidelity simulator for the development and testing of autonomous race cars, (ii) the \LTR~environment which enables rapid prototyping, training, and testing in this simulated environment, and (iii) the \ltr~task which defined dataset characteristics and concrete driving-inspired metrics for evaluation. \ltr~addresses the lack of complex learning environments and introduces the challenging task of simulated, high-performance racing. While human experts have demonstrated strong results on this task, both using the \ltr~framework as well as in competition racing, learning agents have not. We have provided relevant racing metrics and baseline results for classical control, RL, and IL agents as well as human experts, and we are releasing reference implementations and model checkpoints to further advance the research. The \ltr~suite of tasks and metrics will continue to expand in the future including the introduction of multi-agent racing. We hope to someday see agents reach superhuman, real-world performance in autonomous racing.

\section*{Acknowledgements}
There are many people that helped create this task environment. We are grateful to the RoboRace community, in particular, the Hive team which has developed the simulated racetrack maps. We thank developers Bohui Fang, Ignacio Maronna Musetti, Jikai Lu, Zihang Zhang, and Xinnan Du for their support. We also thank Maxim Integrated Products for supporting our efforts. This work was supported, in part, by a fellowship from Bosch Research Pittsburgh.

% JF: We don't have to worry about using \small, here; references do not count toward the content page limit (of 8 pages); we can have ``unlimited" pages for references -- http://iccv2021.thecvf.com/node/4
\bibliographystyle{ieee_fullname}
\bibliography{root}

\clearpage
\appendix

%--------------------------
\section{Racing Simulator Details}

\begin{figure*}
\begin{center}
    \includegraphics[width=\linewidth]{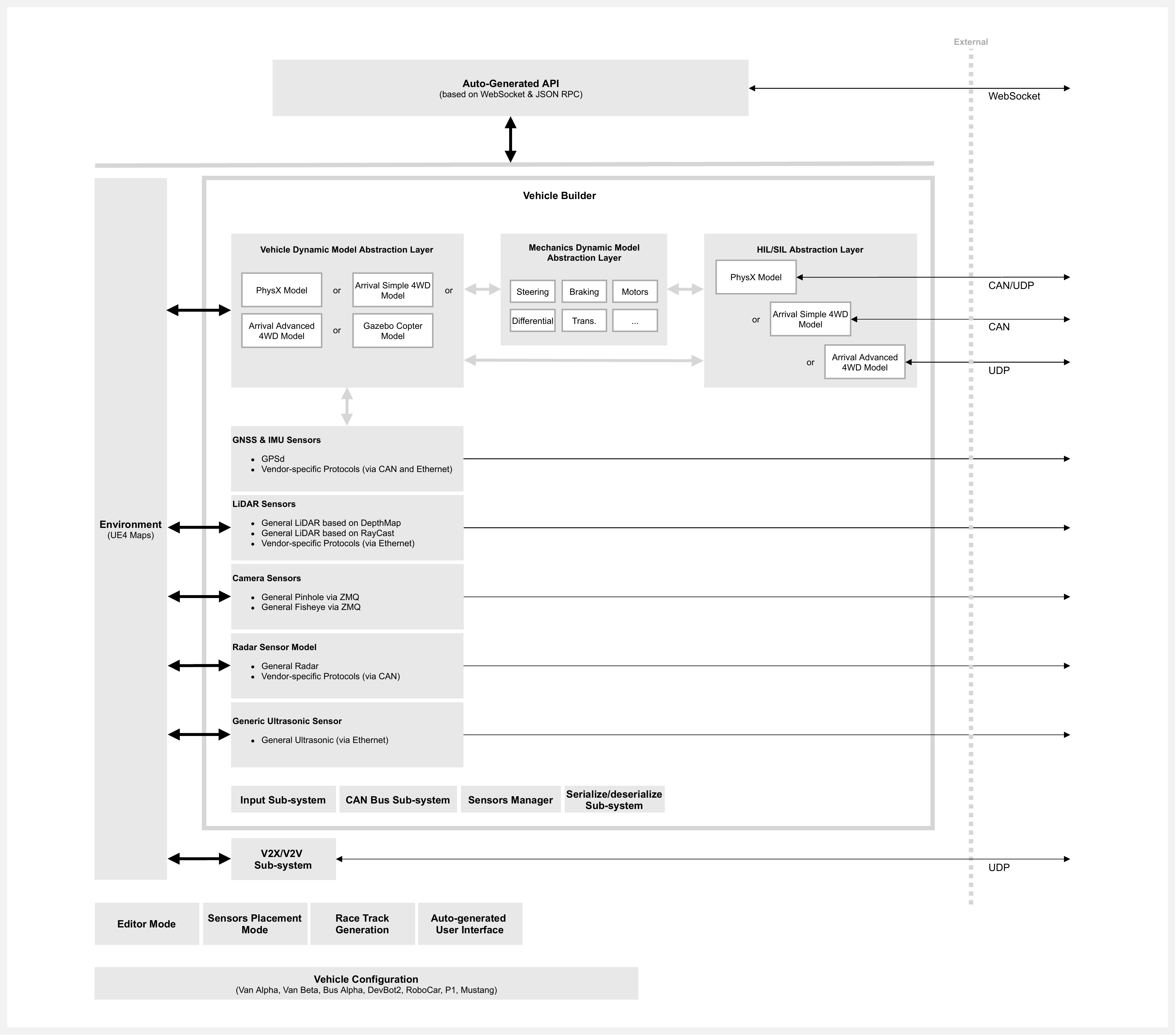}
\end{center}
   \captionsetup{margin=2.5cm}
   \caption{Overview of the racing simulator.}
\label{fig:sim_overview}
\end{figure*}

\subsection{Client and Server Information Exchange}

The agents and the racing simulator act together as a client-server system. The racing simulator includes both a physics and graphics engine and provides numerous communications mechanisms for a variety of use cases. Figure \ref{fig:sim_overview} summarises the simulator system architecture.

\subsubsection{Simulator State}
Management of the simulator's state is done through a web-socket interface, allowing for two-way communication and for clients to update the state of the simulator including the ability to:

\begin{itemize}
    \item Change the map
    \item Change the type of vehicle
    \item Change the pose of the vehicle
    \item Change the input mode
    \item Turn on/off debugging routines
    \item Turn on/off sensors
    \item Modify sensor parameters
    \item Modify vehicle parameters
\end{itemize}

\subsubsection{Simulator-to-Agent Communication}
The simulator communicates to agents primarily with sensory information including:

\begin{itemize}
    \item LiDAR data from 4 independent sensors
    \item RADAR data from the vehicle's radar sensor
    \item Images from the front-facing camera
    \item Pose information from the inertial measurement unit on the vehicle
    \item Additional data about the state of the vehicle such as brake pressure and tire speed \textit{per wheel}
    \item Ground-truth information about other vehicles
    \item Ground-truth information about virtual objects
\end{itemize}

The camera publishes images using Transmission Control Protocol (TCP) while the others publish sensory data using User Data Protocol (UDP) or over a Controller Area Network (CAN). While the \LTR~framework exclusively supports software-in-the-loop simulation, and therefore, only virtual CAN buses, the racing simulator also supports hardware-in-the-loop simulation and physical CAN buses.

\subsubsection{Agent-to-Simulator Communication}
Agents can communicate racing actions to the simulator in a variety of ways:

\begin{itemize}
    \item Via a keyboard or joystick for human drivers
    \item UDP packets with steering, acceleration, and gear requests
    \item Through a safety layer with longitudinal acceleration and curvature requests
    \item Via various API modes which allow for more granular control of the vehicle including individual motor torques and brake pressure requests
\end{itemize}

Consistent with the simulator-to-agent above, agent-to-simulator communication can be done over virtual or physical CAN buses.

%--------------------------
%\subsection{Simulator Racetracks}
\subsection{Additional Visualisation}

\begin{figure*}
\setlength{\tabcolsep}{2pt}
\begin{tabular}{ccc}
  \includegraphics[width=57mm]{assets/thruxton-overhead-1.png} &
  \includegraphics[width=57mm]{assets/lvms-overhead-1.png} &
  \includegraphics[width=57mm]{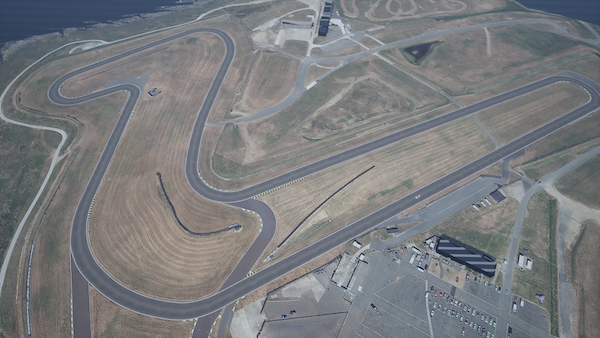} \\
  %\small (a) first & \small (b) second & \small (c) third \\[6pt]
  \includegraphics[width=57mm]{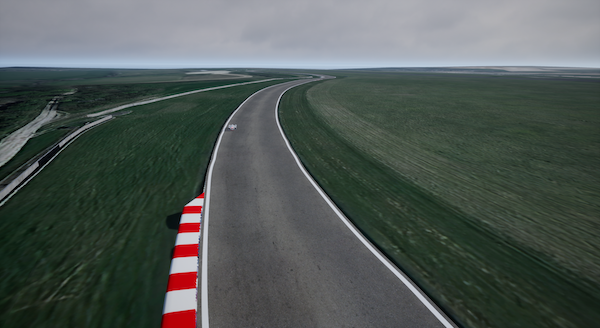} &
  \includegraphics[width=57mm]{assets/lvms-s-curve.png} &
  \includegraphics[width=57mm]{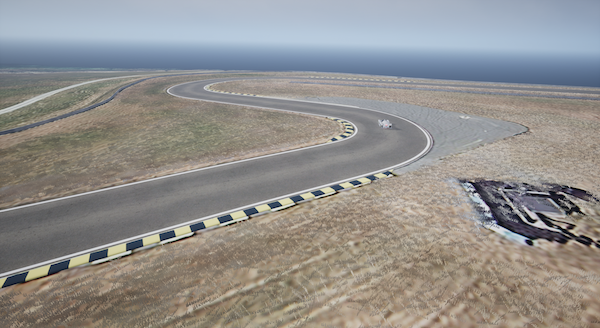} \\
  %\small (a) first & \small (b) second & \small (c) third \\[6pt]
  \includegraphics[width=57mm]{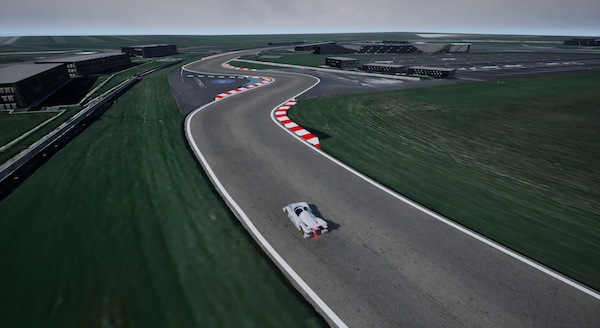} &
  \includegraphics[width=57mm]{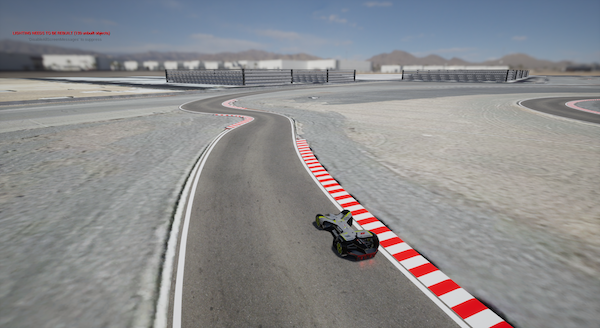} &
  \includegraphics[width=57mm]{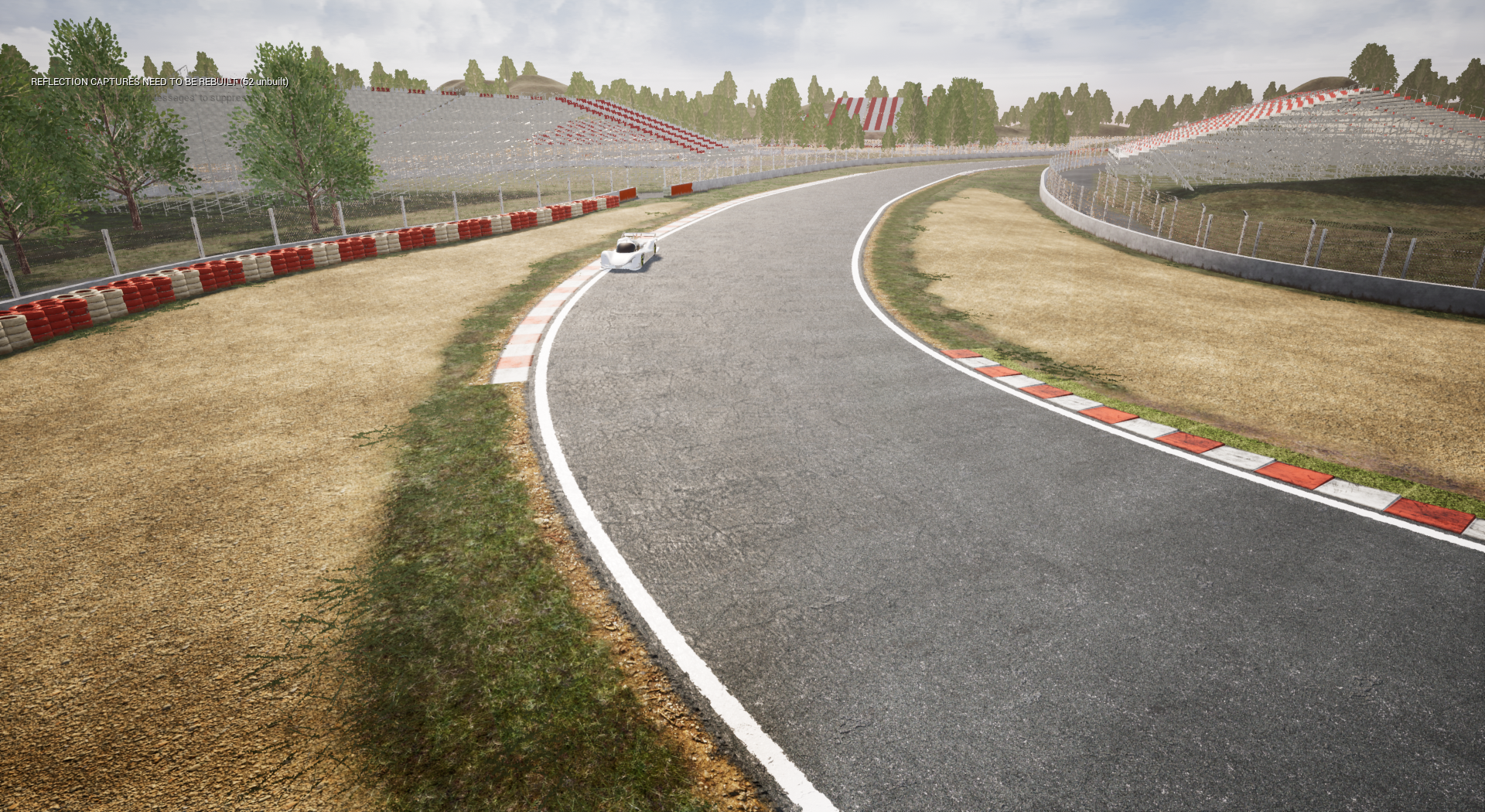}\\
  %\small (a) first & \small (b) second & \small (c) third \\[6pt]
  \includegraphics[width=57mm]{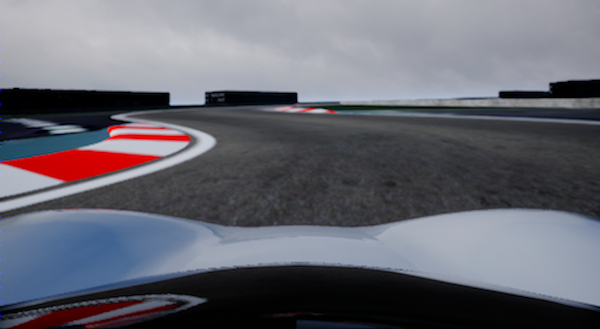} &
  \includegraphics[width=57mm]{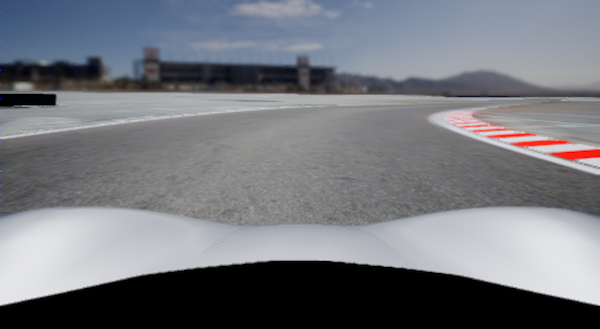} &
  \includegraphics[width=57mm]{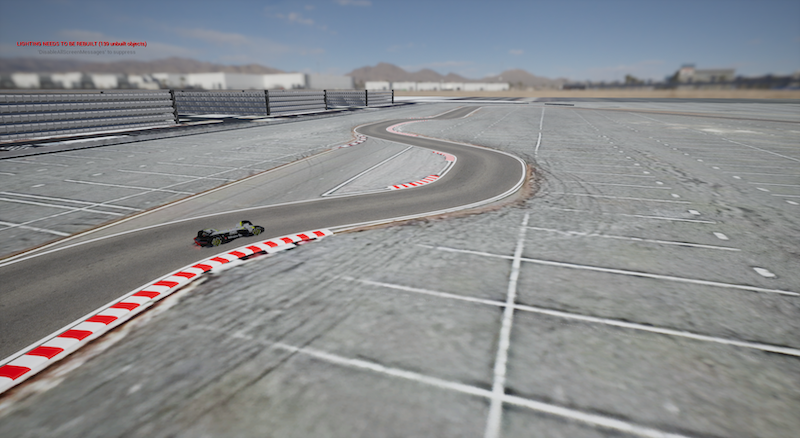}\\
  %\small (a) first & \small (b) second & \small (c) third \\[6pt]
 \includegraphics[width=57mm]{assets/robocar.png} & 
 \includegraphics[width=57mm]{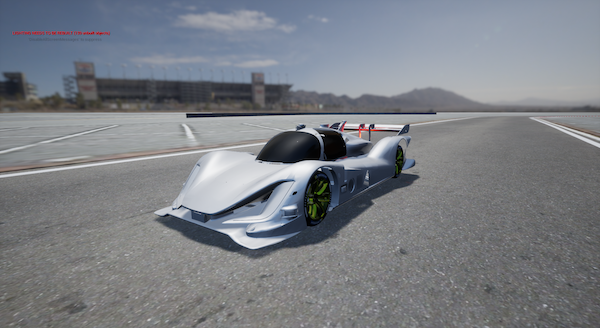} & \includegraphics[width=57mm]{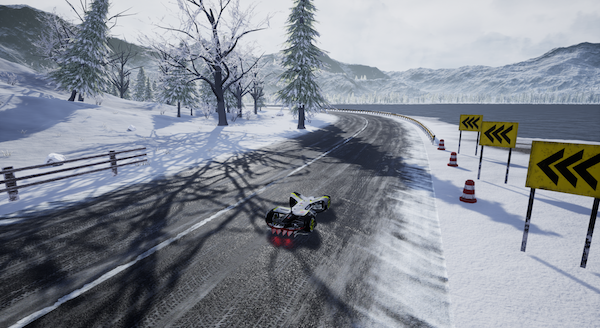}\\
%\multicolumn{3}{c}{\includegraphics[width=57mm]{assets/wintertrack-1.png} }\\
%\multicolumn{3}{c}{(e) fifth}
\end{tabular}
\caption{\small \textit{First column, top four rows}: the Thruxton Circuit race track, United Kingdom, is infamous for its long straightaways, high speeds, and a difficult speed-trap near the finish line. \textit{Second column, top four rows}: the North Road race track at Las Vegas Motor Speedway, United States, includes the sharp turns and merciless speed traps and adds a vision-processing challenge for learning agents, due to the lower contrast between the track and its surroundings. \textit{Third column, top four rows}: the Anglesey Circuit race track, United Kingdom, features two prominent straights and several harrowing turns. \textit{Last row}: the racing simulator features multiple car models, sensor placements, weather conditions, and additional tracks.}
\label{fig:/simulator_racetracks}
\end{figure*}

The racing simulator features multiple real world racetracks each with unique features that challenge human and autonomous agents alike (see Figure \ref{fig:/simulator_racetracks}).

%--------------------------
\section{Dataset Details}

We generate a rich, multimodal dataset of expert demonstrations from the training racetracks (\texttt{Track01:Thruxton} and \texttt{Track02:Anglesey}), in order to facilitate pre-training of agents via, e.g., imitation learning (IL). The \ltr~dataset contains multi-sensory input at a 100-millisecond resolution, in both the observation and action spaces. See Table 1 in the main paper for a complete list of available modalities. The expert demonstrations were collected using a model predictive controller (MPC) that tracks the centerline of the race track at a pre-specified reference speed. Important parameters for this centerline MPC expert included acceleration range of [-1, 1], steering range of [-1, 1], and image H$\times$W dimensions of $384\times512$. This training dataset contains 10,600 samples of each sensory and action dimension, in this first version, which includes 9 complete laps around the track. Demonstrations were saved as invidual step-wise transitions, using \texttt{numpy.savez\_compressed}\footnote{\url{https://numpy.org/doc/stable/reference/generated/numpy.savez_compressed.html}}, with the following as \texttt{dict} fields in the data: (i) \texttt{img} with shape $(384, 512, 3)$; (ii) \texttt{multimodal\_data} with shape $(30,)$; (iii) and \texttt{action} with shape $(2,)$. The fields in \texttt{multimodal\_data} correspond to the vector dimension mappings, indicated in Table \ref{tab:mm_data_dims}.

\begin{table}[h]
\small
  \centering
    \captionsetup{margin=0cm}
    \caption{Vector dimension mappings, to which the data fields in \texttt{multimodal\_data} $(30,)$ correspond.}
  \label{tab:mm_data_dims}
  \begin{tabular}{ll}
    \toprule
    \textbf{Array indices} & \textbf{Description} \\
    \hline
    0 & steering request \\
    1 & gear request \\
    2 & mode \\
    3, 4, 5 & directional velocity in $m/s$ \\
    6, 7, 8 & directional acceleration in $m/s^2$ \\
    9, 10, 11 & directional angular velocity \\
    12, 13, 14 & vehicle yaw, pitch, and roll, respectively \\
    15, 16, 17 & center of vehicle coordinates in $(y, x, z)$ \\
    18, 19, 20, 21 & wheel revolutions per minute (per wheel) \\
    22, 23, 24, 25 & wheel braking (per wheel) \\
    26, 27, 28, 29 & wheel torque (per wheel) \\
    \bottomrule
  \end{tabular}
\end{table}

%\newpage
Future version releases of \ltr~will include access to new simulated tracks (also modelled after real tracks, from around the world) as well as expert traces generated from these additional tracks|across various weather scenarios, in challenging multi-agent settings, and within dangerous obstacle-avoidance conditions. 

\section{Additional Agent Details}

\subsection{RL-SAC Model Details}

\begin{table}[h]
  \centering
    \captionsetup{margin=1.2cm}
    \caption{\texttt{RL-SAC} model hyperparameters}
  \label{tab:sac_hyperparams}
  \begin{tabular}{ll}
    \toprule
    \textbf{Hyperparameter} & \textbf{Value} \\
    \hline
    Buffer size & 100,000 \\
    Gamma & 0.99 \\
    Polyak & 0.995 \\
    Learning rate & 0.001 \\
    Alpha & 0.2 \\
    Batch size & 256 \\ 
    Start steps & 1000 \\
    Learning steps & 5 \\
    \bottomrule
  \end{tabular}
\end{table}

\noindent The \texttt{RL-SAC} agent learns from image embeddings rather than raw pixels. The encoder used is a convolutional variational autoencoder (VAE) which was trained prior to, and frozen during, the \texttt{RL-SAC} agent learning. The VAE was trained to encode RGB images of with a width and height of 144 pixels each into a latent space of size 32. The encoder architecture consisted of 4 convolutional layers, each followed by a ReLu activation, with a kernel size and stride of 4 and 2, respectively. The result of the convolutions was passed through a single fully connected layer to the compressed representation. Binary cross entropy loss and an Adam optimiser were used for training.

The \texttt{RL-SAC} agent was trained for 1,000 episodes which was approximately 1 million steps in the environment. We trained this agent in vision-only mode, so it only had access to the camera's images. The agent passed the encoded images through two fully connected layers with 64 units each and a final layer with an output shape of 2, matching the environment's action space. Gradient updates were taken at the conclusion of episodes, and the training hyperparameters are listed in Table~\ref{tab:sac_hyperparams}.

\subsection{MPC Agent Details}
The MPC problem is summarised by Equation \ref{eq:mpc_problem}. The objective (Equation \ref{eq:mpc_objective}) is to minimise the tracking error with respect to a reference trajectory, in this case the centerline of the race track at a pre-specified reference speed, with regularisation on actuations, over a planning horizon of T time steps. $\mathbf{Q}$ and $\mathbf{R}$ are both diagonal matrices corresponding to cost weights for tracking reference states and regularising actions. At the same time, the MPC respects the system dynamics of the vehicle (Equation \ref{eq:mpc_sys}), and allowable action range (Equation \ref{eq:mpc_action}). 

\begin{subequations}\label{eq:mpc_problem}
\small
    \begin{align}
    \min_{\mathbf{a}_{1:T}}  \quad & \sum_{t=1}^{T}\left[(\mathbf{s}_t-\mathbf{s}_{ref, i})^T \mathbf{Q}(\mathbf{s}_i-\mathbf{s}_{ref, i}) +\mathbf{a}_i^T \mathbf{R} \mathbf{a}_i\right] \label{eq:mpc_objective}\\
    \textrm{s.t.} \quad & \mathbf{s}_{t+1} = f(\mathbf{s}_t, \mathbf{a}_t), \quad \forall t = 1, \dots, T \label{eq:mpc_sys}\\
    & \underline{\mathbf{a}}\leq \mathbf{a}_t\leq \bar{\mathbf{a}}\label{eq:mpc_action} 
    \end{align}
\end{subequations}

Specifically, we characterise the vehicle with the kinematic bike model\footnote{This set of equations is defined with respect to the back axle of the vehicle and is used for generating expert demonstrations. The kinematic bike model defined with respect to the centre of the vehicle is also included in our code base.} \cite{kong2015kinematic} given in Equation \ref{eq:bike_model}, where the state is $\mathbf{s}=[x, y, v, \phi]$, and the action is $\mathbf{a}=[a, \delta]$. $x, y$ are the vehicle location in local east, north, up (ENU) coordinates, $v$ is the vehicle speed, and $\phi$ is the yaw angle (measured anti-clockwise from the local east-axis). $a$ is the acceleration, and $\delta$ is the steering angle at the front axle. 
\begin{subequations}\label{eq:bike_model}
\begin{align}
\dot{x}&=v\cos(\phi)\\
\dot{y}&=v\sin(\phi)\\
\dot{v}&=a\\
\dot{\phi}&=v\tan{\delta}/L
\end{align}
\end{subequations}

A key challenge is that the ground truth vehicle parameters were not known to us. Aside from $L$ defined as the distance between the front and rear axle, the kinematic bike model expects actions, i.e. acceleration and steering, in physical units, while the environment expects commands in $[-1, 1]$. The mapping is unknown to us, and non-linear based on our observations. For instance, acceleration command = 1 results in smaller acceleration at higher speed. In the current implementation, we make a simplifying assumption that $a=k_1\times$ acceleration command, and $\delta=k_2\times$ steering command.  

We use the iterative linear quadratic regulator (iLQR) proposed in \cite{li2004iterative}, which iteratively linearizes the non-linear dynamics (Equation \ref{eq:bike_model}) along the current estimate of trajectory, solves a linear quadratic regulator problem based on the linearized dynamics, and repeats the process until convergence. Specifically, we used the implementation for iLQR from \cite{amos2018differentiable}. The parameters used by the MPC are summarised in Table \ref{tab:mpc_hyperparams}. 

\begin{table}[h]
  \centering
    \captionsetup{margin=1.2cm}
    \caption{MPC  parameters}
  \label{tab:mpc_hyperparams}
  \begin{tabular}{ll}
    \toprule
    \textbf{Parameter} & \textbf{Value} \\
    \hline
    $\mathbf{Q}$ & $\text{diag}([1, 1, 1, 16])$ \\
    $\mathbf{R}$ & $\text{diag}([0.1, 1])$ \\
    $v_{ref}$ & 12.5 m/s\\
    $\mathbf{\bar{a}}$ & $[1, 0.2]$ \\
    $\mathbf{\underline{a}}$ & $[-1, -0.2]$ \\
    $L$ & 2.7 m \\
    $k_1$ & 10 \\
    $k_2$ & 6 \\
    $T$ & 6 \\
    \bottomrule
  \end{tabular}
\end{table}

%--------------------------

\section{Metric Equations}

We quantify the parametric curvature of a trajectory in Eqn. \ref{eq:curvature}, with $x_{t}'$ denoting $\frac{dx}{dt}$ at time $t$, and we summarise the curvature of the entire path as $\kappa_{rms}$ in Eqn. \ref{eq:curvature_rms}:

\begin{equation}
    % numerical estimate of curvature
    % curvature
    \kappa_{t} = \frac{x_{t}'y_{t}''-y_{t}'x_{t}''}{\Big((x_{t}')^2+(y_{t}')^2\Big)^\frac{3}{2}}
    \label{eq:curvature}
\end{equation}

\begin{equation}
    % Summarize curvature as the RMS value
    \kappa_{rms} = \sqrt{\frac{1}{T}\left(\sum_{t=0}^{T} \kappa_{t}^{2}\right)}
    \label{eq:curvature_rms}
\end{equation}

\noindent We measure \textit{Trajectory efficiency} as $\rho$ in Eqn. \ref{eq:traj_efficiency} based on the curvature, $\kappa_{rms}$, of the race track and the racing agent's trajectory.

\begin{equation}
    % Trajectory efficiency
    \rho = \frac{\kappa_{rms,\:racetrack}}{\kappa_{rms,\:trajectory}}
    \label{eq:traj_efficiency}
\end{equation}

\end{document}